\documentclass[conference]{IEEEtran}
\IEEEoverridecommandlockouts
\usepackage[]{amsmath}
\usepackage{amssymb}
\usepackage{multirow}
\usepackage{verbatim}
\usepackage{graphicx}
\graphicspath{ {images/} }
\usepackage{subcaption}
\usepackage{epstopdf}
\usepackage{tabularx}
\usepackage{booktabs} 
\usepackage{multirow}
\usepackage{arydshln} 
\usepackage{url}
\usepackage{hyperref} 
\usepackage{textcomp} 
\usepackage{float} 
\usepackage{xcolor}
\usepackage{setspace}
\usepackage{wrapfig}
\usepackage{mathtools, cuted}

\newcommand{\linebreakand}{%
  \end{@IEEEauthorhalign}
  \hfill\mbox{}\par
  \mbox{}\hfill\begin{@IEEEauthorhalign}
}

\title{Learning Robust Policies for Generalized Debris Capture with an Automated Tether-Net System}

\author{
Chen Zeng\IEEEauthorrefmark{1}
,
Grant Hecht\IEEEauthorrefmark{1}
,
Prajit KrisshnaKumar\IEEEauthorrefmark{1}
\thanks{\IEEEauthorrefmark{1} Ph.D. Student, Department of Mechanical and Aerospace Engineering}
,
Raj K. Shah\IEEEauthorrefmark{2}
\thanks{\IEEEauthorrefmark{2} M.S. Student, Mechanical and Aerospace Engineering}
,
Souma Chowdhury\IEEEauthorrefmark{3}
\thanks{\IEEEauthorrefmark{3} Associate Professor, Mechanical and Aerospace Engineering,
Corr. author email: soumacho@buffalo.edu}
,
Eleonora M. Botta\IEEEauthorrefmark{4}
\thanks{\IEEEauthorrefmark{4} Assistant Professor, Mechanical and Aerospace Engineering}\\
\IEEEauthorblockA{\normalsize\itshape
University at Buffalo, Buffalo, NY, 14260}
}

\newcommand\blfootnote[1]{%
  \begingroup
  \renewcommand\thefootnote{}\footnote{#1}%
  \addtocounter{footnote}{-1}%
  \endgroup
}

\begin{document}
\maketitle
\begin{abstract}
Tether-net launched from a chaser spacecraft provides a promising method to capture and dispose off large space debris in orbit. This tether-net system is subject to several sources of uncertainty in sensing and actuation that affect the performance of its net launch and closing control. Earlier reliability based optimization approaches to design control actions however remain challenging and computationally prohibitive to generalize over varying launch scenarios and target (debris) state relative to chaser. To search for a general and reliable control policy, this paper presents a reinforcement learning framework that integrates a proximal policy optimization (PPO2) approach with net dynamics simulations. The latter allows evaluating the episodes of net-based target capture, and estimate the capture quality index that serves as the reward feedback to PPO2. Here, the learnt policy is designed to model the timing of the net closing action based on the state of the moving net and the target, under any given launch scenario. A stochastic state transition model is considered in order to incorporate synthetic uncertainties in state estimation and launch actuation. Along with notable reward improvement during training, the trained policy demonstrates capture performance (over a wide range of launch/target scenarios) that is close to that obtained with reliability based optimization run over an individual scenario. 
\blfootnote{**This work was supported by the
NSF award CMMI 2128578. Any opinions, findings, conclusions, or recommendations expressed in this paper are those of the authors and do not
necessarily reflect the views of the NSF.}

~
\end{abstract}

\section{Introduction}\label{sec1}

Active Debris Removal (ADR) using tether-net systems has been proposed and studied as a promising solution to the space debris problem \cite{guang2012space, wormnes2013throw, benvenuto2015dynamics, botta2018deployment,hausmann2015ohb, benvenuto2012implementation, bombelli2012multidisciplinary}. Among others, Botta et al. \cite{botta2016evaluation,botta2017energy,botta2019simulation} conducted extensive research on the dynamics of the deployment and capture phases of net-based debris removal missions. 

The majority of existing works are based on the assumptions of capturing a specific target with ideal launch conditions, with a handful of pioneering works (Salvi \cite{salvi2014flexible}, Botta et al. \cite{botta2017energy,botta2016evaluation}, and Endo et al. \cite{endo2020study}) looking into the robustness of deployment or capture under various net launch conditions in the absence of a closing mechanism. In this type of mission, it is crucial to guarantee a successful capture of the target in the presence of uncertainties, and to ensure that the target remains wrapped by the net when the chaser tugs it to its disposal orbit. These uncertainties can be attributed to measurement errors, inaccuracies in net launch control, and to the estimation of the debris and chaser vehicle inertial and attitude states. However, to date, very little work has been performed to study the effects of uncertainties on the system robustness, especially in the presence of a closing mechanism, with the exception of work by Chen et al. \cite{Chen2022analysis}.   

The tether-net capture system presents a multidisciplinary robust design-control problem with pertinent constraints under uncertainties. In our preceding research\cite{shah2021reliability}, a reliability-based design optimization process was proposed to optimize the launching and/or closure of the net under the influences of uncertainties for a known fixed debris target. This process models uncertainties and performs Bayesian optimization to determine launch strategies that maximize the capture success rate. However, such a method only searches for a single strategy that applies to a predefined debris status, meanwhile requiring considerable computing power. Therefore, while it provided valuable insight on robust debris capture under uncertainty, this approach is ultimately not suitable as a flexible and flight-worthy solution. 

\begin{figure*}[htbp]
    \centering
    \includegraphics[width=0.8\linewidth]{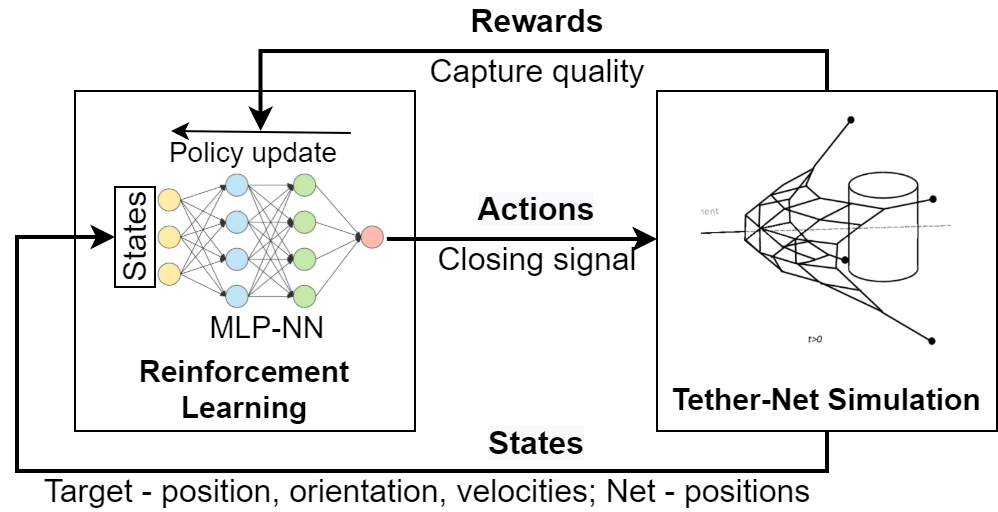}
    \caption{Proposed Policy Learning Process for the Tether-Net System}
    \label{fig:concept1}
\end{figure*}

Artificial Neural Networks (ANN) are playing an emerging role as decision-support models in various intelligent autonomous systems \cite{dounis2009advanced}. As a universal function approximator, an ANN is capable of mapping states to actions in autonomous systems, a.k.a. a policy model. Various ANN fitting (learning) methods have seen demonstrations on robotics and control applications. Popular learning methods include Reinforcement Learning\cite{baxter1999knightcap,peters2003reinforcement}, Supervised Learning\cite{caruana2006empirical}, Imitation Learning\cite{hussein2017imitation}, Neuroevolution\cite{risi2015neuroevolution,hansen2003reducing}, etc., among which the advanced reinforcement learning\cite{schulman2017proximal} and neuroevolution\cite{behjat2019adaptive} methods are directly applicable to launching and wrapping control of tether-net systems. These aforementioned machine learning methods bring capabilities of adapting to system uncertainties, and selecting optimal actions (policies) according to various debris characteristics. Computation-heavy reliability analyses required in the optimizations can also be reduced, retaining the computing load similar to that of reliability-based optimizations. 

This paper proposes a machine-learning-based policy optimization of a one-shot robotics system with environment adaptability and robustness under uncertainties. The case study features the launching and wrapping process for a tether-net space debris capture system. The machine learning framework adapts to various environmental parameters (including net geometry and states of the chaser and the target), considers multiple sources of uncertainties (including inaccuracies of state estimation, errors in launching and wrapping parameters, and sensing errors), and determines the optimal 
wrapping parameters that maximize the probability of a successful capture. Environment parameters are relative to the capture system (e.g., the mass of the corner masses and the geometry of the net) and the state of the debris (e.g., the distance to the chaser, its orientation, and its motion). 
The wrapping policy comprises only triggering of the closing mechanism. The capture success is evaluated by the Capture Quality Index, or CQI (interpreted later in the manuscript). 
A case study scenario is presented with standalone wrapping policy learning with programmed launching employing a state-of-the-art learning technique, Proximal Policy Optimization 2 (PPO2)\cite{schulman2017proximal}.


The remainder of this paper is organized as follows: In Section \ref{sec2}, the architecture of the simulator is presented briefly, together with the models implemented for the different components of the system. In Section \ref{sec3}, the adopted machine-learning-based policy optimization is discussed. Section \ref{sec4}  presents validation of the optimization results, and Section \ref{sec5} concludes the paper with a discussion of results and limitations of the work, and associated planned future work.

\section{Modeling: Tether-Net Launching and Wrapping Mechanics} \label{sec2}




Inherited from the preceding work\cite{botta2019simulation,BottaPhdThesis,shah2021reliability}, the modeled system comprises of a chaser carrying a square-shaped net with 4 corner masses and a closing mechanism around the perimeter. The tether-net system is simulated in Vortex Studio, a powerful multibody dynamics simulation platform designed for real-time simulation of complex mechanics.
The net is modeled with the standard lumped-parameter approach. The mass properties of the net are lumped into small spherical rigid bodies placed at the physical knots of the net, herein referred to as nodes. The axial stiffness and damping properties of the net's threads are represented with massless springs and dampers in parallel between the nodes. 

The chaser spacecraft is modeled as a cubic rigid body in the simulation. The main tether, linking the net to the chaser, is modeled with a series of slender rigid bodies, modeled as relaxed prismatic joints to simulate the axial and bending stiffness and damping properties.

\begin{figure*}[h]
\centering
\includegraphics[width=0.7\linewidth]{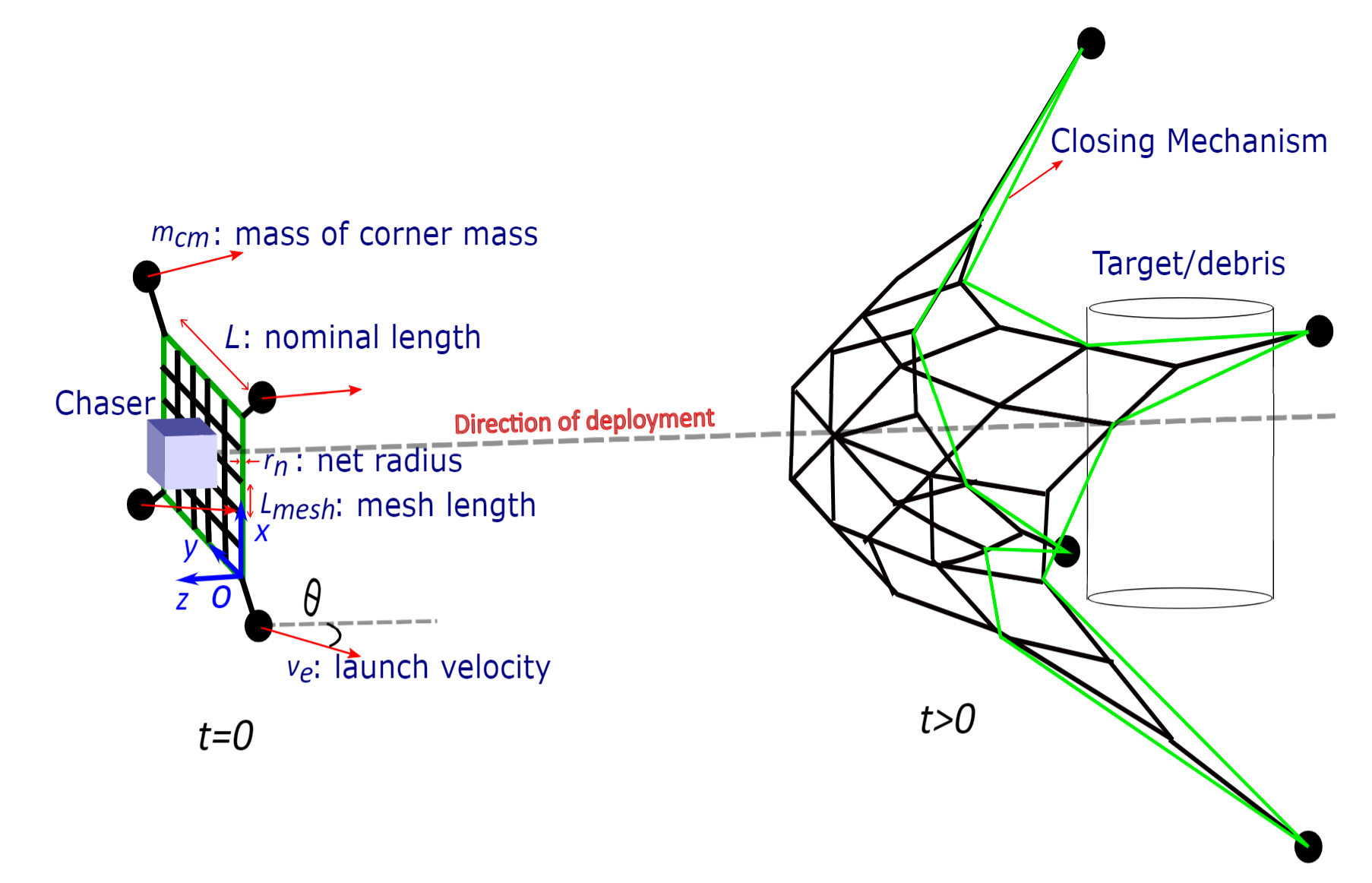}
\caption{Sketch of the Modeled Tether-net System}
\label{fig:concept2}\vspace{-0.2cm}
\end{figure*}

The closing mechanism applies a drawstring interlaced with the perimeter of the net, as shown in Fig. \ref{fig:concept2}. The drawstring passes through 8 nodes on the perimeter as well as the 4 corner masses, and is winched independently from the main tether\cite{botta2019simulation}.
When the closing mechanism is activated, constant forces are made to act between each pair of adjacent nodes along the drawstring, pulling the nodes together until contact. Upon contact, the node pairs are locked to keep the mouth of the net closed for the rest of the maneuver. 

The detailed design of the tether-net system is fixed as the derived result from preceding work\cite{shah2021reliability}. Table \ref{tab:hardware} lists the design parameters.

\begin{table}
	\begin{center}
		\caption{Design and properties of the tether-net system}
		\label{tab:hardware}
		\begin{tabular}{lll}
			\toprule
			Variable & Description & Value \\
			\midrule
			$L$ & Side Length of Net & 22.0 m \\
			$L_{\texttt{mesh}}$ & Side Length of Mesh & 1.375 m \\
            $r_{\texttt{n}}$ & Thread Radius & 6 mm \\
            $m_{\texttt{n}}$ & Total Mass of Net & 100 kg \\
            $m_c$ & Mass of Corner Mass & 10.0 kg \\
			\hdashline
			$\mathbf{V}_{\texttt{eb}}$ & Launching Velocity & $[3.30, 3.54, 7.16]$ m/s \\
			\bottomrule
		\end{tabular}
	\end{center}
\end{table}

\section{Learning the Optimal Launching and Closing Policies}\label{sec3}

\subsection{Defining the learning task} \label{sec3a}

Reinforcement learning models the actions as Markov Decision Processes \cite{bellman1957markovian} (MDP) that comprise a comprehensive state space with a relatively compact action space. There are 2 sets of actions (launching and closing) taking effect on different stages of the capture operation, we are focusing on closing within the scope of this paper. 

The policy model of the closing signal is time variant, while the action itself is a logical variable (boolean). 
The observations (state space) are assumed to be estimated by employing readily available sensors, including: 1) internal measurement units (IMUs), cameras, or Lidar mounted on the chaser vehicle, 2) IMUs and cameras attached to the corner masses, as well as 3) monitoring from the Earth. Therefore, the state space mainly consists of the position, orientation, and velocities of the target, as well as the positions of the corner masses of the net. In addition, the velocity of launching the corner masses, the simulation time, and a flag indicating the actuation of closing are also part of the observation. 
Some parameters, like the position of the net's center of mass, are obtained from the simulation for calculations of the reward and the constraints, but excluded from the state space since they are unattainable with the conceived sensors. Table \ref{tab:states} lists the state space and the action space parameters.

\begin{table*}[]
    \centering
    \caption{Parameters of State and Action Spaces}
    \label{tab:states}
    \begin{tabular}{llll}
    \toprule
        Type & Variable & Data Type & Boundaries \\
        \midrule
        \multirow{7}{*}{State}& Time & Scalar & $0$ to $120$ s \\
        & Target Coordinates ($p_{\texttt{target}}$) &  Cartesian & $[-10, -10, 0]$ to $[10, 10, 50]$ m\\
        & Target Orientation &  Euler angles & $[0, 0, 0]$ to $[2 \pi, 2 \pi, 2 \pi]$ rad\\
        & Target Angular Velocity & Euler angles & $[-1, -1, -1]$ to $[1, 1, 1]$ rad/s\\
        & Pos. of Corner Masses (4) & Cartesian & $[-22, -22, 0]$ to $[22, 22, 72]$ m\\
        & Closure Flag ($f_c$) & Boolean & - \\   
        & Launching Velocity &  Cartesian & $[1, 1, 1]$ to $[5, 5, 10]$ m/s\\
        \midrule
        Action & Closing Signal &  Boolean & $0$ and $1$ \\
    \bottomrule    
    \end{tabular}
\end{table*}

The objective of reinforcement learning is to train the policy model to find the optimal timing of sending the closing signal, based on observations regarding the states of the net and the target. The policy model adapts a reward function defined as: 
\begin{strip}
\begin{subequations}
\label{eq:formulation}
\begin{align}
\underset{\mathbf{Q}}{\text{max:}} \quad & R_A= 
    \left( \sum_{t_0}^{t_{\texttt{close}}} r(t) + r_\texttt{end} \right) / n_{\texttt{steps}}
\\
\text{where:} \quad & r(t) = 
\begin{cases}
w_1 \cdot \left(C_1-\left|\frac{V_{\texttt{n}}-V_{\texttt{t}}}{V_{\texttt{t}}}\right|\right) + w_2 \cdot \left(C_2-\left|\frac{S_{\texttt{n}}-S_{\texttt{t}}}{S_{\texttt{t}}}\right|\right) + w_3 \cdot \left(C_3 -\left|\frac{q_{\texttt{n}}}{q_{\texttt{t}}}\right|\right) + w_4 \cdot (N_{\texttt{L}}-C_4) + \dots \\ 
\hspace{9mm}  0.4\cdot(\min(t,t_1)-4.6) - (0.12\cdot(\max(t,t_2)-t_2))^2 + C_5  \quad \text{(no premature closing)} \\
-(t_1 - t_{\texttt{close}})^2 \hspace{70.5mm} \text{(with premature closing)}
\end{cases}
\\
& r_\texttt{end} = 
\begin{cases}
    w'_1 \cdot \left(C'_1-\left|\frac{V'_{\texttt{n}}-V_{\texttt{t}}}{V_{\texttt{t}}}\right|\right) + w'_2 \cdot \left(C'_2-\left|\frac{S'_{\texttt{n}}-S_{\texttt{t}}}{S_{\texttt{t}}}\right|\right) + w'_3 \cdot \left(C'_3 -\left|\frac{q'_{\texttt{n}}}{q_{\texttt{t}}}\right|\right) + \dots \\ 
    \hspace{55.5mm} w'_4 \cdot (N'_{\texttt{L}}-C'_4) + C'_5 \quad \text{(closing started before 60 s)} \\
    C_6 \hspace{96mm} \text{(not closing by 60 s)}
\end{cases}
\end{align}
\end{subequations}
\end{strip}

In which $\mathbf{Q}$ represents the policy model; $R_A$ represents the episodic (mean) reward; $t$ represents the simulation time; $t_0$ stands for the initial and final time steps; $t_{\texttt{close}}$ represents the time when the closing signal is issued; $t_1$ and $t_2$ are manually configured time triggers; $n_{\texttt{steps}}$ stands for the number of (learning) steps in the episode; $r(t)$ represents the reward at time step $t$; $r_\texttt{end}$ represents the bonus end-of-episode reward; $V_{\texttt{n}}$, $S_{\texttt{n}}$, and $q_{\texttt{n}}$ stand for the enclosed volume, surface area, and the center-of-mass position of the net at time $t$; $V'_{\texttt{n}}$, $S'_{\texttt{n}}$, and $q'_{\texttt{n}}$ stand for those at time ($t_{\texttt{close}}+20$ seconds); $V_{\texttt{t}}$, $S_{\texttt{t}}$, and $q_{\texttt{t}}$ stand for the volume, surface area, and the center-of-mass position of the target; $N_{\texttt{L}}$ is the number of locked node-pairs around the edge (12 pairs in total); $C_1$ through $C_6$, as well as $w_1$ through $w_4$ are tuning weights, in addition to $C'_1$ through $C'_5$ and $w'_1$ through $w'_4$; $f_C$ refers to the closure flag. 

A premature closing is defined as:
\begin{equation}
    \label{eq:closing}
    \begin{cases}
    \text{\textbf{True} } & \text{if }(f_C=1) \cap (t_{\texttt{close}}<15 \text{ s}) \cap \dots \\
     & [(q_{\texttt{n}}(t_{\texttt{close}})>12 \text{ m}) \cup (|\overline{p}_{\texttt{cm}}-p_{\texttt{target}}|>10\text{ m})] \\
    
    \text{\textbf{False} } & \text{otherwise}
    \end{cases}
\end{equation}
where $\overline{p}_{\texttt{cm}}$ represents the mean position of the corner masses, and $p_{\texttt{target}}$ represents the position of the target.

A crucial part of the reward is largely dependent on the distances between the net and the target in terms of position, surface area, and volume. Such formulation is inspired by the Capture Quality Index (CQI)\cite{barnes2020quality}, which estimates the successfulness of a capture with the aforementioned measurements. The number of locked node-pairs ($N_{\texttt{L}}$) indicates the possibility of the target slipping out of the net, and is also considered in the formulation for added robustness.

The conditional formulations in the reward function ensures some substantial penalties for premature closing (too far away from the target) or delayed closing (too late), and a substantial bonus ($C'_5$) for appropriately-timed closing. Timing is vital for a successful capture, but the time-variant nature of the closing signal makes it difficult for the policy model to learn to avoid premature action. A reward function solely based on the CQI is insufficient to distinguish between an obvious failure and a possibly successful scenario, hence the added penalty that overrules the reward is necessary to avoid possible exploitation of the CQI formulation. Since the reward function returns positive rewards through the majority of an episode, the policy model could exploit the formulation by not closing at all. The bonus reward ($C'_5$) for well-timed closing and the penalty ($C_6$) for not closing at all would work together to prevent the policy model from refusing to close throughout the episode.

We train the policy model in stages controlled by step count, while adjusting the tuning weights as the stages progress. The CQI part of the end-of-episode reward $r_{\texttt{end}}$ is gradually magnified as the policy model learns to avoid premature closing. Penalty for delayed closing ($C_6$) is inactive until the policy model learns to refuse to close. Table \ref{tab:constants} lists the values of the adjustable parameters applied through various stages of training.
\begin{table*}[]
    \centering
    \caption{Coefficients for the Reward Formulation}
    \label{tab:constants}
    \begin{tabular}{llllllllllll}
        \toprule
         \multirow{2}{*}{Unchanged} & $w_1$ & $w_2$ & $w_3$ & $w_4$ & $C_1$ & $C_2$ & $C_3$ & $C_4$ & $C_5$ & $t_1$ & $t_2$ \\
         & 0.025 & 0.025 & 0.2 & 0.125 & 3 & 3 & 5 & 2 & 2 & 15 & 20 \\
         \midrule
         \multirow{2}{*}{} & $w'_1$ & $w'_2$ & $w'_3$ & $w'_4$ & $C'_1$ & $C'_2$ & $C'_3$ & $C'_4$ & $C'_5$ & $C_6$ \\
         \hdashline
         0 to 66,000 steps & 0.05 & 0.05 & 0.4 & 0.125 & 3 & 4 & 6 & 0 & 50 & 0 \\
         66,001 to 300,000 steps & 0.1 & 0.1 & 0.8 & 0.25 & 3 & 3 & 6 & 0 & 50 & 0 \\
         300,001 to 800,000 steps & 1 & 1 & 8 & 2.5 & 3 & 3 & 6 & 0 & 50 & -50 \\
         800,001 to 1,500,000 steps & 2 & 2 & 16 & 5 & 3 & 3 & 3 & 2 & 100 & -50 \\
         \bottomrule
    \end{tabular}
\end{table*}

Two scenarios were conceived: 1) learning a standalone wrapping policy while launching is programmed; 2) learning launching and wrapping policies simultaneously. Scenario 2 calls for training two policy models in one simulation, which brings in questions like reward crediting \cite{nguyen2018credit} and the orders of execution. Limited by time and computing constraints, the case study in this paper is confined to scenario 1). 
The programmed launching policy for scenario 1 was presented in a previous work \cite{shah2021reliability} (values listed in Table \ref{tab:hardware}). 


\subsection{Design of experiments and uncertainty modeling}

The Design of Experiments samples a range of the target's distance, initial orientation, and initial rotating angular velocity. Every episode of training is sampled with random choices of the forementioned initial states of the target. Table \ref{tab:doe} lists the variations of DoE.

\begin{table}[]
    \centering
    \caption{Initial States of the Target}
    \label{tab:doe}
    \begin{tabular}{lll}
        \toprule
        Name & Minimum & Maximum \\
        \midrule
        Dist. to chaser & 25 m & 35 m \\
        Orientation & $[0, 0, 0]$ rad & $[\frac{\pi}{2}, 0, 0]$ rad \\
        Angular Vel. & $[0, -\frac{\pi}{18}, -\frac{\pi}{18}]$ rad/s & $[0, \frac{\pi}{18}, \frac{\pi}{18}]$ rad/s \\
        \bottomrule
    \end{tabular}
\end{table}

The Simulation cannot directly simulate the onboard sensors and actuators in high fidelity, therefore uncertainties are modeled in sensing, state estimation, and actuation of the system by applying stochastic noises to the parameters. Sources of uncertainties include: 1) estimated state of the debris target and the corner masses of the net, 2) velocity of launching the net, 3) timing of triggering the closing mechanism, and uniquely 4) soft dynamics of the net. The uncertainty of the net dynamics is aleatoric and unavoidable, while the majority of the uncertainties in estimation and control are epistemic \cite{hora1996aleatory}. 

\begin{figure*}[htb]
\centering
\includegraphics[width=0.8\linewidth]{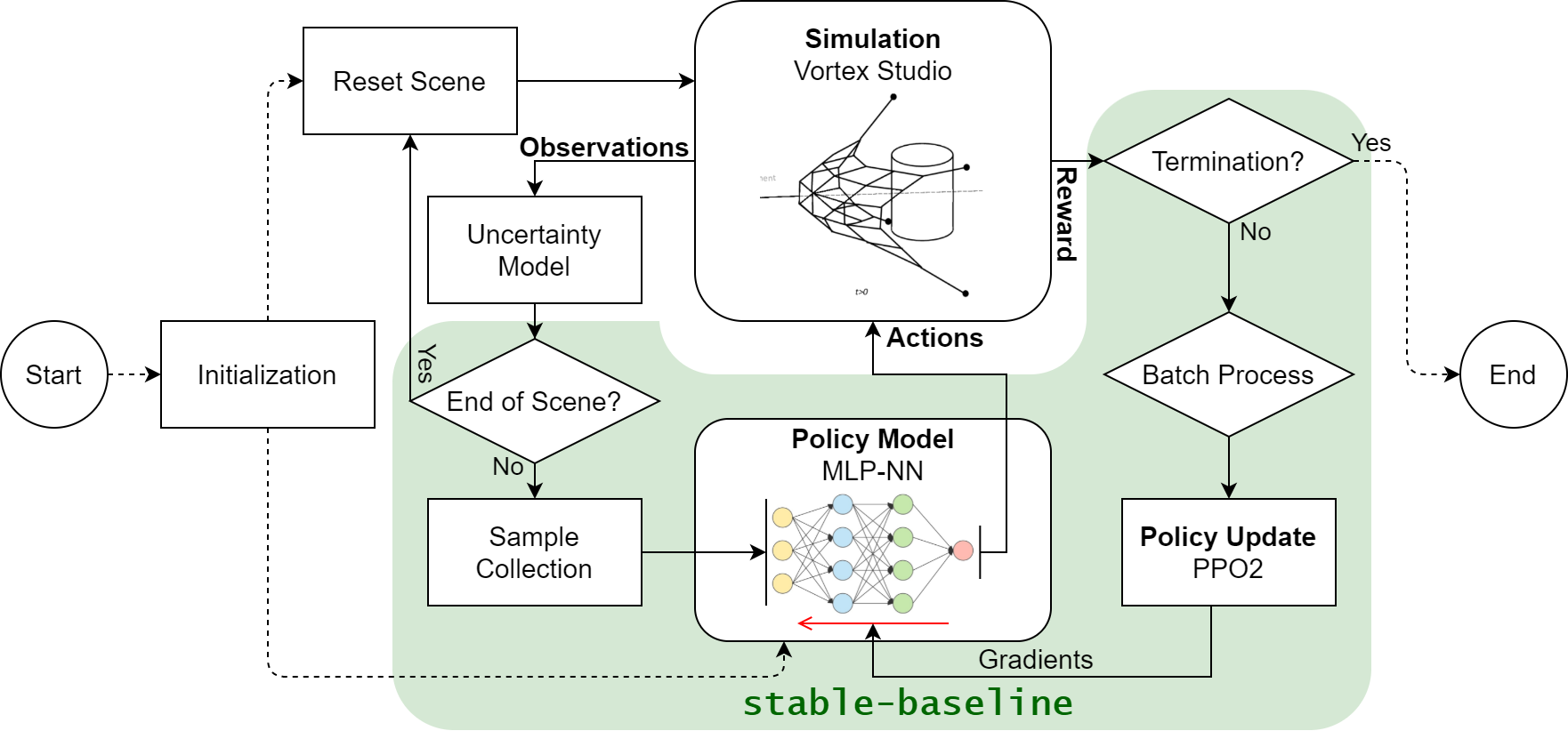}
\caption{The Workflow of Reinforcement Learning on the Tether-Net Capture Mission}
\label{fg:diagram}
\end{figure*}
Gaussian noises modeled after Table \ref{tb:noises} are applied upon the observed parameters from the simulation before delivery to the learning algorithm. Noises of the constant parameters (target orientation and launching velocity) are one-shot, while those of the continuously monitored parameters are sampled at every time step of learning.

\begin{table*}
	\begin{center}
		\caption{Noise Levels of Modeled Uncertainties}
		\label{tb:noises}
		\begin{tabular}{llll}
			\toprule
			Sampled During: & Noise Source & Data Type & Margin of Error ($2\sigma$) \\
			\midrule
			\multirow{3}{*}{Step-wise} & Target Orientation & 3D Vector & $\pm [\pi/36, \pi/36, \pi/36]$ rad\\
			 &  Target Angular Velocity & 3D Vector & $\pm [\pi/36, \pi/36, \pi/36]$ rad/s\\
			 & Corner Masses Position & 3D Vector & $\pm [0.1, 0.1, 0.25]$ m \\
			\hdashline
			\multirow{2}{*}{Only Once} & Target Position (CoM) & 3D Vector & $\pm [0.1, 0.1, 0.25]$ m \\
			 & Launching Velocity & 3D Vector & $\pm [0.05, 0.05, 0.1]$ m/s \\
			\bottomrule
		\end{tabular}
	\end{center}
\end{table*}


\subsection{Learning and validation techniques}

Herein, the learning technique known as Proximal Policy Optimization 2 (PPO2) from stable baselines \cite{stable-baselines} is applied within a case study to provide an opportunity to evaluate and compare advanced reinforcement learning with prior results obtained by robust optimization under a fixed environment \cite{shah2021reliability}. PPO2 is a state-of-the-art actor-critic reinforcement learning method which has demonstrated high efficiency, wide adaptability, and robust reliability \cite{DBLP:journals/corr/abs-1811-02553}. 

PP02 employs a gradient update based on the experiences collected on the mini-batch of interactions with the environment. Upon update completion, the experiences collected are then discarded and the next update is based on new experiences. PPO2 has been proven to demonstrate less variance during the training process when compared to alternative learning techniques which ensures a smoother training process \cite{DBLP:journals/corr/abs-1709-06560}. A multi-layer perceptron model consisting of 2 layers with 64 neurons is used within the deep Q network. This network is trained for 500 episodes and the trained policy is then validated for 100 episodes. Parameters used for the learning process are provided in Table \ref{tab:PPO2_params}.

\begin{table}[htbp]
\small
\centering
\caption{Reinforcement Learning Parameters}
\begin{tabular}[t]{cc}
    \toprule
         Algorithm & PPO2  \\
         
         \midrule
         Neural Network type & Multi-Layer Perceptron   \\
         Total Timesteps & 1,500,000  \\
         Learning Rate & 2.5e-4 \\
         Discount Factor & 0.999 \\
         Number of Steps & 128 \\
         Entropy Coefficient & 0.01 \\
        Clip Range & 0.2 \\
         Value Function Coefficient & 0.5 \\
         Max. Gradient Norm & 0.5 \\
    \bottomrule
\end{tabular}
\label{tab:PPO2_params}
\end{table}%


For the final evaluation, a reliability sampling process was performed to examine the probability of success for the trained policy model. This involved a Monte Carlo sampling process to sample the impact of the uncertainties, in which $N = 100$ is number of independent simulations executed with the trained policy model. The evaluated probabilities of success are then compared to the success rate achieved by robust optimization under a fixed environment\cite{shah2021reliability}. 

\section{Results} \label{sec4}

The learning trials were executed on a Windows workstation with 16 CPU cores with parallel computing with 30 workers for the episode evaluations. The learning process progresses in 20,000-to-500,000-step stages as we examine the learning rate and adjust the reward weights in between stages. By the conclusion of this paper, a total of 1.5 million steps of learning have been finished. The time cost of learning was 41.2 hours.

\begin{figure}[htb]
    \centering
    \includegraphics[width=0.9\linewidth]{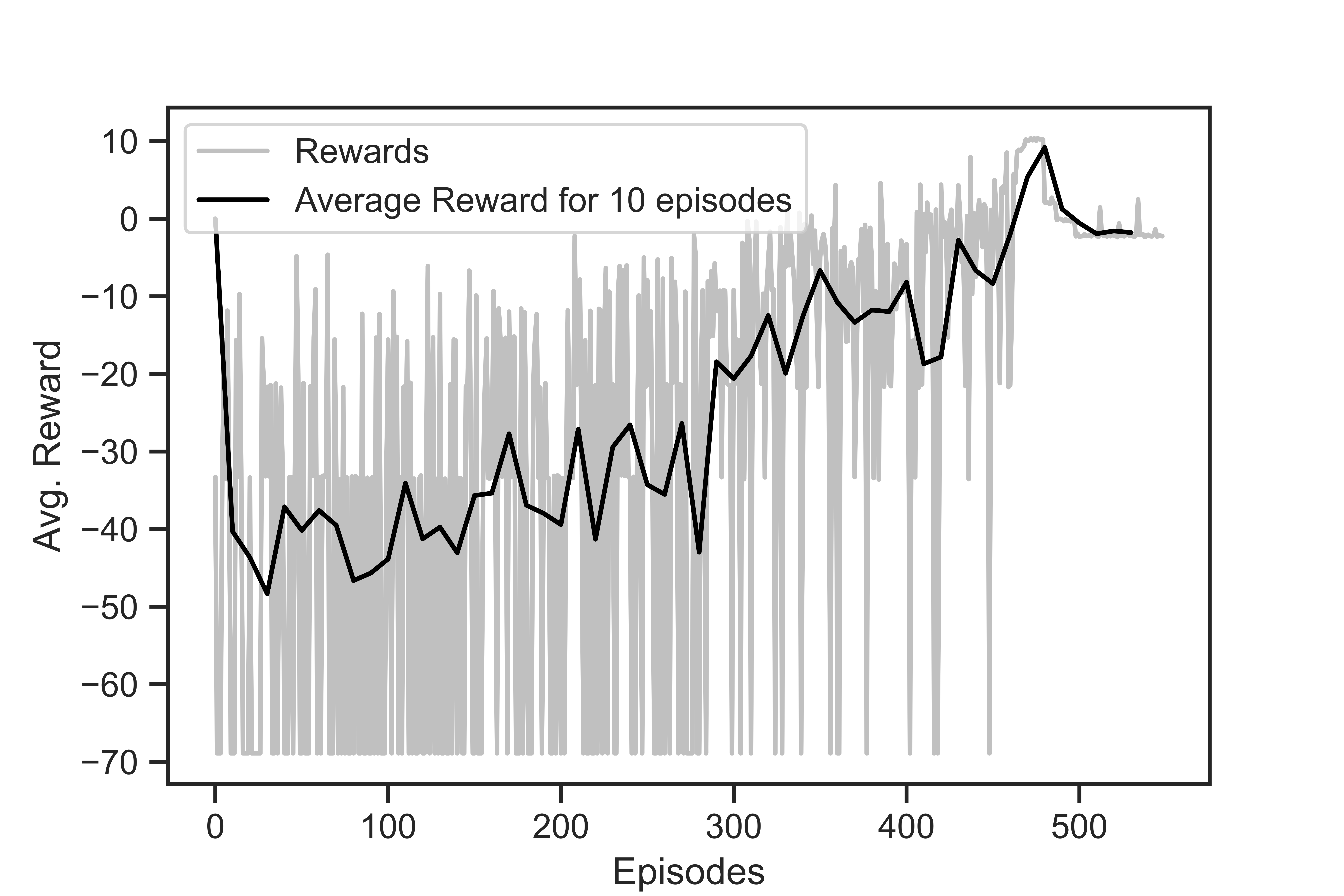}
    \caption{History of Episodic Rewards}
    \label{fg:convhist}
\end{figure}

The history of episodic reward from one of the 30 workers is displayed in Figure \ref{fg:convhist}. This specific worker finished 549 episodes in 46,437 steps. Since the episode tends to extend longer as the learning progresses, the later 89\% of the steps only contributed to 80 episodes. The lowest episodic reward is -69.0, and the highest episodic reward is 9.7. 

The episodic reward plot shows strong fluctuations, indicating the learning rate is unstable throughout the learning process and has yet to approach convergence. The 10-episode mean reward is also shown within the figure, which provides a more stable indication of the range and the trend of the reward values. The policy model received negative rewards throughout most of the episodes, but just managed to receive near-zero or positive rewards after 460 episodes. The trend of the rewards and the fluctuations are both signs of insufficient training. The episodic mean reward is unsuitable as a direct measurement of capture quality (for the successfully closed cases), considering the lengths of episodes fluctuate in a wide margin.

To evaluate the quality of the trained policy models, we tested all the neural networks obtained via different stages of learning with a Monte Carlo sample set sampled from the same initial states as shown in Table \ref{tab:doe} and applied the same level of uncertainties as shown in Table \ref{tb:noises}. The best-performing policy model so far is from the end of 120,000 steps, which is compared to results from the previous work. The CQI values of the test result are calculated and compared with an optimized closing time under a static initial condition obtained from the predeceasing paper\cite{shah2021reliability} (only closing was optimized for a single initial state). The mean CQI value of the policy model tests is 1.035, and the percentage of the samples with CQI values lower than 2 (seen as secure captures) is 94\%. In contrast, the optimized fixed close timing managed to achieve 96\% success rate and a mean CQI of 1.010 for the single initial state. The policy model of closing achieves a high success rate in a range of initial states, near that for the fixed close timing optimized for a single initial state, suggesting the policy model is approaching the maximal possible reliability. The policy model should keep improving if given more learning steps, and we expect the converged policy model to outperform the optimization results, especially in situations with large deviations.


\section{Concluding Remarks} \label{sec5}
A machine-learning-based tether-net system wrapping policy optimization for ADR  with environment adaptability and robustness under uncertainties was proposed and a trade study involving standalone wrapping policy learning with programmed launching employing a state-of-the-art learning technique, Proximal Policy Optimization 2 (PPO2), was performed. 

Despite cutting off the learning process early, evaluation of the policy learning results shows that the proposed approach for wrapping policy learning proves promising, resulting in capture reliability comparable to earlier robust Bayesian optimization which involved a similar computational load, despite optimizing the wrapping strategy under a much wider range of state scenarios with larger uncertainty. 

Future work will investigate standalone launching policy learning with programmed wrapping, and simultaneous learning of launching and wrapping policies. Additional machine learning tools, including reinforcement learning algorithms and advanced neruoevolution techniques\cite{behjat2019adaptive} will also be applied for experiments.


\bibliographystyle{aiaa}
\bibliography{biblist}

\end{document}